# Normativity and Productivism: Ableist Intelligence? A Degrowth Analysis of AI Sign Language Translation Tools for Deaf People

Nina Seron[1], Poppy Fynes[1],

*University of York*[1]

# Table of Contents

*Nina Seron, Poppy Fynes - SAINTS CDT - january 2026*

**INTRODUCTION**

'Sign' languages, of any geographical or accentual variation, understandably face continuous scrutiny under the ever-present popularity of verbal dictation and the accommodating of audism. Through this, many potential problems arise with the current lack of accessible communication for those who rely on such sign languages for essential conversation - exacerbated by the rise of 'AI' systems.

Such AI systems regularly take the form of recognition and interpretation models, designed to provide seamless and accurate translation, like sign apse and sign avatar. In reality these systems are built from biased data and created without any input from Deaf communities. Such models are widely used and accepted by their hearing counterparts - who remain ignorant to the inherent culture, semantics and colloquial language present in such deep and meaningful gestural language systems.

The promises of these AI tools lie on the assumption that AI will be able to fully replicate human sign language communication, which ignores the tension between the complexity of sign language communication and the rationalizing needed for developing these tools. The promises of these AI systems function on the other assumption that more technological innovation could solve the problems of the social realm, and could alleviate the structural injustice deaf people face. But that illusion of accessibility might not translate into real inclusivity for deaf people, in the sense that this minimal overcoming of physical barriers with AI might also have the potential to threaten the conditions for authentic social progress for deaf people, which should be the end goal of inclusivity.

The question is : who are the AI tools for sign language translation really benefitting ?

The UN's *Disability-Inclusive Language Guidelines* define inclusivity as "removing barriers"[1] – but the radical inclusivity that we aspire to in this paper should transcend the conventional framing around "equal resources and opportunities" for the disabled, and aim to effectively remove all barriers and determinisms, in realizing autonomy, i.e. the freedom of an individual to govern oneself (and thus act) according to one's own principles, values, and reasoning[2]. We will assess the results in inclusivity of these tools with regard to autonomy, because it is pivotal to political ecology philosophy, as a condition for human emancipation and flourishing, and thus a morally desirable political goal for deaf and able-bodied people.

We argue following Jacques Ellul and other degrowth philosophers (Illich, Gorz) that these AI tools for sign language translation, like any technique before them, will shape deaf people's environment, subjectivity, experience and consciousness according to economic interests and goals of growth and productivity in such a way that these tools won't serve deaf people but perpetuate their alienation.

---

[1] United Nations Office at Geneva. *Disability-Inclusive Language Guidelines*. Geneva: United Nations Office at Geneva, 2019.
https://www.ungeneva.org/sites/default/files/2021-01/Disability-Inclusive-Language-Guidelines.pdf

[2] Sarah Buss, "Personal Autonomy," in *Stanford Encyclopedia of Philosophy*, ed. Edward N. Zalta (Stanford, CA: Metaphysics Research Lab, Center for the Study of Language and Information, Stanford University, Spring 2019), accessed January 7, 2026,
https://plato.stanford.edu/entries/personal-autonomy/

In this paper we are interested in investigating if AI tools designed for inclusivity, communication and accessibility could not in fact defeat their own purposes by growing their opposite instead. That would unveil a contradiction/paradox by which technical solutions for inclusivity, could in fact be per essence and *de facto* ableist. Ableism has been defined as *'ideas, practices, institutions, and social relations that presume able-bodiedness, and by so doing, construct persons with disabilities as marginalized… and largely invisible "others."'*[3] This philosophical problem matters because the strong and popular adhesion to technosolutionism might serve a depoliticization in which both collective responsibility of hearing people and the potential for emancipatory action of deaf people are annihilated.

In the first section, we support the Ellulian analysis of the subversion of the promethean myth : Technique is no longer our product, serving us, or a means to an end. It is now our environment and has the power to shape our experience and subjectivity as much as we shape it.

In the second section, we explain how artificial "intelligence" through code, datafication, rationalization, mathematization, in fact operates on a reductionist model of human intelligence, language, and communication.

In the third section, we show how that reductionism is synonymous with alienation of subjectivity resulting from the AI tool's standardization, homogenization and uniformization of language/communication that impoverishes individual consciousness and culture.

In the fourth section, we explore the counter-productive (Illich) aspects of the AI tool in the possible psychological harms of social alienation.

In the fifth section, we highlight how these harms collide with capitalist interests and dynamics both embedded in and powered by Technique.

In the last section, we explore paths for the reappropriation of language by deaf people, to serve autonomy, justice, emancipation, i.e real "inclusivity".

---

[3]Vera Chouinard, "Making Space for Disabling Differences: Challenging Ableist Geographies," *Environment and Planning D: Society and Space* 15, no. 4 (1997): 380.

1. **AI tools in the technical system as our "milieu"**

Technique as depicted in the traditional promethean myth, used to be a mediation between man and its environment. But Ellul observed in *The Technological System* and *Technological bluff* that Technique has become a system : it's even the mediation between humans, as seen in the use of the telephone, the radio and TV[4], and today, social media. It now orders the social world and for we are social animals, Technique shifts from instrumental tools born of culture to our whole environment or "milieu".

In the same way, language used to serve the higher function of a mediating structure between humans and their environment, as a *poietic,* fulfilling and emancipating factor that helped to make sense of the world, create meaning, and relate to each other socially.

The case of AI translation tools for deaf people is a prime example of a further conquest and invasion of Technique over language, entrenching its hegemony as the determining factor of society: communication ceases to be intersubjective meaning-making, it is a depersonalised and transactional false interaction, made through the machine and not with the other. Moreover, it is a totalizing factor : it aims to account for the whole of reality, experience, or meaning that was previously born of the social and the vernacular. Former culture is replaced by our new artificial yet natural order : the technical system as our milieu, with its language where *"nature, poetry, the magical, the mythical and the symbolic"* disappear.[5] What can't be explained in technical terms is neutralized.[6] Technique is indeed now our environment, for it shapes at the forefront lived experience and the social sphere.

Rather than encountering others through reciprocal linguistic adaptations, deaf individuals must encounter the world through interfaces that impose their own norms of intelligibility. For instance, AI when used for the recognition of signed language in the absence of human interpretation imposes new norms such as the change in the *formality* of the language. This is, in the sense that those words and phrases which hold a more colloquial meaning, or those that are present due to accentual differences are now changed, or considered 'incorrect' in some technical form. The environment here ceases to be a shared space of meaning making and becomes what we may see as a technically organised field, demanding the sanitisation and self-adjustment of a rich and diverse language. In this sense we may argue our deaf subject to be not merely mediated by technique but disciplined and configured by it.

---

[4]Jacques Ellul, *Le système technicien* (Paris: Calmann-Lévy, 1977), p43
[5]Jacques Ellul, *Le système technicien* (Paris: Calmann-Lévy, 1977), p40
[6] Jacques Ellul, *Le système technicien* (Paris: Calmann-Lévy, 1977) p51

## 2. <u>Technique operates on a reductionist normativity</u>

Intelligence is not reduced to algebra, and Ellul posits that the computer can't be made truly intelligent, for it has no imagination and no spontaneity, no dreams, fears or desires, characteristics considered by Ellul as constitutive of intelligence, for they are the provokers of thought. The intelligence of the program is merely an imitation and never an accurate rendering of the human world.[7] But AI seeks to simulate the intellectual operations of Man, eventually to replace it, constituting progressively what Illich called a "radical monopoly". With "radical monopoly", submission to Technique becomes mandatory : we have no choice but to adapt to the tool and technical imperatives instead of the other way round – or we risk marginalization, exclusion, unemployment... Technique therefore aims in its essence to condition Man, to adapt to its project of limitless economic expansion.

But the process of what we call "rational" thinking that is imitated, is a temporal product, as studied by cognitive sciences and in phenomenology.[8] *"But human memories and predictions are not the same as those of a computer: they are memories of joys, achievements and failures, predictions mixed with fears and hopes."*[9] That means there's nothing such as a purely "rational" thought for humans, and while the computer poses as perfectly and purely rational –though it operates on particular postulates encoded that may escape reason– Technique can't fully capture that human experience ; because intelligence, thought, reasoning, are not a storing or accumulation of data, but the result of social relationships, "accidents and joys", the result of projects, obsessions, intuitions and intentions.

The datafication of sign language must obey some rules of rationalization for Technique to act and interfere : social relationships have to be rationalized, leaving out affects and sentimentality in the process. Technique models communication on the modalities of Technique, rendering it not intersubjective communication but communication of information, under the form of quantifiable data. [10] However, communication is a social process of conveying meaning between agents that involves more than data transfer : it needs intent, interpretation, and shared context, whereas technical informational exchange doesn't require shared understanding: one party can send information without the other interpreting it socially.

But this end put to the social, "natural", and *poetic* is ultimately a reduction of the complexity of what phenomenology with Husserl calls our "lifeworld" *(Lebenswelt)*. The concept is also used by André Gorz (following Sartre), for whom the defense of "nature" at the heart of political ecology means the defense of that "*world as immediately or directly*

---

[7] Jacques Ellul, *Le bluff technologique* (Paris: Hachette, 1988). p203
[8] Li S, Wang Z, Sun Y. Relationship between Thinking Dispositions, Working Memory, and Critical Thinking Ability in Adolescents: A Longitudinal Cross-Lagged Analysis. J Intell. 2024 May 21;12(6):52. doi: 10.3390/jintelligence12060052. PMID: 38921687; PMCID: PMC11204695.
[9] Jacques Ellul, *Le bluff technologique* (Paris: Hachette, 1988).p199
[10] S. Li, Z. Wang, and Y. Sun, "Relationship between Thinking Dispositions, Working Memory, and Critical Thinking Ability in Adolescents: A Longitudinal Cross-Lagged Analysis," *Journal of Intelligence* 12, no. 6 (May 21, 2024): 52, https://doi.org/10.3390/jintelligence12060052

*experienced in the subjectivity of everyday life [through] individual, social, perceptual, and practical experiences.*"[11]. The struggle of political ecology is also a struggle against the destruction of ways of life invented by humans, in touch with their needs, their environment and other humans.

Indeed, Technique has to be *"simplistic, reductive, operational, instrumental and reordering"*. That means diversity in traditions, cultures, and social structures are eroded in favor of technical "efficiency" (productivism) and standardization.[12] AI is thus reductionist, and in mediating communication, it is Man that will in turn be standardized and objectified in conformity with the disembodied laws of Technique and its economic and moral goals[13].

In the case of our Deaf subject we consider the datafication of Sign language as a divergence from the true nature of its cultural and communicative roots. Sign language as we have previously argued is a language such that there is distinct linguistic diversity and inherent meaning similar to those languages which we find verbally communicated in almost any society. Unlike such verbalised communication, signed communication extracts meaning through the coordination of body, gaze, facial grammar, rhythm and even shared culture. If we are to generalise such a diverse language to the needs and desires of the machine, then Deaf intelligence is forced to appear as a sequence of isolated gestures, stripped of its embodied, temporal, and imaginative dimensions. The nature of Technique's informational exchange dictates that loss of meaning, since it doesn't aim for shared understanding but merely transmission of information. The deaf person can be understood badly, wrongly, or partially, with no consequence.  Language becomes a sequence of symbols to be decoded, rather than a living medium of thought, culture, and social connection.

Through this process technical rationality is imposed as the standard against which Deaf expression is measured. What is spontaneous, metaphorical or context dependent is rendered unintelligible, while what is regular, repetitive and predictable is thus valorised. The richness of *Deaf cognition*; its use of metaphor role shifting and spatial narration is now reduced to operational features. This confirms Ellul's claims that technique does not merely describe reality but ultimately reorganises it according to its own logic.  In this way, Man does not merely use the machine; he is reformatted by it and the world he inhabits is narrowed to what our machine may recognise and optimise for.

---

[11] *"Life-world," Encyclopædia Britannica*, accessed January 7, 2026, https://www.britannica.com/topic/life-world
[12] Jacques Ellul, *La technique ou l'enjeu du siècle* (Paris: Economica, 2008).P360
[13] Jacques Ellul, *La technique ou l'enjeu du siècle* (Paris: Economica, 2008). P401

### 3.   Colonization of consciousness and alienation of subjectivity

According to Maurice Merleau-Ponty, consciousness is shaped through human encounter, through the experience of communicative acts and the linguistic bonding between humans[14] and that claim has been backed multiple times by sociology[15], anthropology[16], and psychology[17]. Therefore, rather than expanding the lifeworld of the Deaf subject, Technique narrows it, dissolving the thickness of the embodied meaning of intercorporeality[18] into the thin legibility of data and machine interaction.

We argue in this section that the AI tools, in modeling human communication according to its reductionist postulates and processes of informational exchange, risk not only the impoverishment of language and communication but the restructuring of consciousness itself, narrowing the horizons of thought, imagination, subjectivity, memory, knowledge and expression available to Deaf people.

André Gorz writes in The Immaterial, following Husserl, *that mathematization or "thought freed from the body"* brackets *"all the ways of thinking and self-evident facts that are not indispensable to the technique of calculation, including of course, the needs, desires, pleasures, pains, fears or hopes that form the perpetually rewoven fabric of consciousness. The intellect detached from affective life in this way, whose only intention is to function in accordance with the laws and rules of calculation—regarded as the laws and rules of thought freed from irrationality-then discovers layers of reality inaccessible to experience and to other modes of thinking"*[19]

Indeed, Technique requires uniformization because it can't possibly operate on, or adapt to the novel, the moving and the unstable that characterizes variety in the social world. Technical rationality requires and enforces a specific ontology of the immutable  – and thus of the status quo. It lacks complexity, as per the previous section, but it also forbids complexity, and doing so creates a new norm for Man. With "mathematical language", AI structures and determines what it means to be human : causation, time, space, identity, necessity, and possibility, and can influence how we metaphysically represent to ourselves the physical, linguistic, biological, social, political, symbolic, affective. For example, knowing that the AI can't grasp some nuances, and with continuous self-censoring, one can interiorize a form of "efficiency" in signing, leading to a more functionalist language (describing

---

[14] Maurice Merleau-Ponty, *Phenomenology of Perception*, trans. Colin Smith (London: Routledge & Kegan Paul, 1962).
[15] Erving Goffman, *The Presentation of Self in Everyday Life* (Garden City, NY: Doubleday, 1959).
[16] Michael Tomasello, *The Cultural Origins of Human Cognition* (Cambridge, MA: Harvard University Press, 1999).
[17] L. S. Vygotsky, *Mind in Society: The Development of Higher Psychological Processes*, ed. Michael Cole, Vera John-Steiner, Sylvia Scribner, and Ellen Souberman (Cambridge, MA: Harvard University Press, 1978).
[18] S. Tanaka, "Intercorporeality as a Theory of Social Cognition," *Theory & Psychology* 25, no. 4 (August 2015): 455–72, https://doi.org/10.1177/0959354315583035
[19] André Gorz, *The Immaterial*, trans. Chris Turner (London: Seagull Books, 2010). p160

objects, people, institutions through their role or utility) and thereby to thought ; With less abstraction, it's the very process of thinking that could be impaired; and with repetition and integration, it's the social categories (functions) that can be naturalized, and no longer ideologically questioned[20]

That results in cultural anthropisation, because statistics encourage thinking in terms of averages, biases and fewer nuances : everywhere where technical growth exists, it *"has the same causes, produces the same effects, provides people with a similar living environment, imposes a certain type of work on them, involves the same changes in social and political organisations, and requires the same conditions for its growth and development... And this is true regardless of historical origins, geographical situations or possibilities, or social or political regimes.".* [21]We can hypothesise that the growth of technical mediation in sign language communication will also produce standardization, homogenization and uniformization in subjectivity, destroying what we call culture : for culture is about roots, diversity and specificity in language that come from personal experience and history. With AI softwares for sign language, culture (as experience, intersubjectivity and interrelatedness, constitutive of consciousness) is rendered "obsolete", as per Ellul's prophecy.[22] Following André Gorz's hypotheses[23], one may argue that this colonization of consciousness impairs and predates on the very same raw material necessary for developing autonomy.

---

[20]Herbert Marcuse, *One-Dimensional Man* (London: Routledge, 2007)p149
[21]Jacques Ellul, *Le système technicien* (Paris: Calmann-Lévy, 1977), p166
[22]Jacques Ellul, *Le bluff technologique* (Paris: Hachette, 1988). p81
[23] André Gorz et al., "La conquête de l'autonomie," *Autogestions* NS no. 8–9 (1982): 187–204, https://www.persee.fr/doc/autog_0249-2563_1982_num_15_8_1485

### 4. Counter-productivity : how the "inclusive" tool could grow exclusion

In the previous section, we pointed out how the lifeworld of the deaf person may stagnate or decline. One can fear that impoverished and alienated consciousness/subjectivity could be mislabelled as the lack of expressive, emotional, social, or linguistic intelligence or maturity, and other personal failings to be blamed on the deaf person, that could potentially pave the way to forms of isolation and marginalization.

What is certain is that the institutional, private and public assumptions that automated translation will be sufficient or satisfactory, assumptions that drive the growth of AI mediation for deaf people, annihilate the moral and social responsibility of the State and the able-bodied collective for learning, teaching, and maintaining sign language. Accessibility and inclusivity are depoliticized. Techniques thus will function as an alibi: communication is no longer something that must be collectively negotiated but something outsourced to a system whose neutrality is presumed and whose limits are rendered invisible.This leads such groups to become dependent on the machine for intelligibility rather than to live and experience life as part of a shared, human linguistic and cultural world. This internalises the feelings of displacement that Deaf people and communities already experience in the wider world.

According to Ivan Illich, in the history of technological innovation, when a tool reaches the threshold of radical monopoly – which is very likely for these AI tools that are substitutive, i.e. introduced not alongside "convivial" human interpretation or sign language learning, but instead of them – an unavoidable consequence is alienation from others. Community is ever more weakened and replaced by technical massification that obliterates the conditions for fulfilling human relationships. Alienation from others is one of the many aspects of what Illich calls the "counter-productivity" of technological development : just as cars created more distance than they have bridged[24], not only will the "inclusive" tool perpetuate the structural injustice of exclusion, it might aggravate it. Ellul shares Illich's analysis :  continuous technological development iatrogenically creates more of the problems it was meant to solve.

Thus, these AI tools could *"exacerbate social maladjustment and backwardness, increase marginalisation; mechanisms designed to give greater freedom result in maximum inevitability; the acceleration of changes in the system leads to a worsening of crises. The multiplication of means leads to the disappearance of ends. The growth of universal power increases social impotence, and the means of power of each individual behave like prostheses that suppress the natural uses of functions."*. [25]

---

[24]Ivan Illich, *Tools for Conviviality* (London: Fontana/Collins, 1973). p21
[25]Jacques Ellul, *Le bluff technologique* (Paris: Hachette, 1988). p206

Ellul notes insightfully that alienation from others born of Technique's moral individualistic project of massification is often externalised in neurosis[26], or what we call "psychological harms" in safety science. And indeed, today social alienation (in isolation, exclusion) is proved to impair decision-making, attention, memory,[27] and to be linked to depression, anxiety, and cognitive decline[28].

Authentic inclusivity through autonomy and for the realization of human flourishing at the heart of political ecology philosophy, is not possible without fulfilling the innate cognitive and emotional need for interrelatedness that is disrupted by what Illich calls "heteronomous" tools (i.e. domineering). On the contrary, making our tools "convivial" would lead to fostering social cooperation.

We could add that social interrelatedness could also play a role in the development of class consciousness, via the shared experience of exploitation and alienation, which would allow the conditions for emancipation from economic constraints that we will tackle in the next section, with a more ecosocialist (thus marxist) approach.

---

[26] Jacques Ellul, *La technique ou l'enjeu du siècle* (Paris: Economica, 2008). P268

[27]. Xu, L. Qiao, S. Qi, Z. Li, L. Diao, L. Fan, L. Zhang, and D. Yang, "Social Exclusion Weakens Storage Capacity and Attentional Filtering Ability in Visual Working Memory," *Social Cognitive and Affective Neuroscience* 13, no. 1 (January 1, 2018): 92–101, https://doi.org/10.1093/scan/nsx139

[28] Roy F. Baumeister, "The Need to Belong: Desire for Interpersonal Attachments as a Fundamental Human Motivation," *Psychological Bulletin* 117, no. 3 (1995): 497–529, https://doi.org/10.1037/0033-2909.117.3.497

## 5. The AI tool as the expression and reproduction of the material project of productivism

During Ellul's time, linguistic research already reduced language to structures, functions and mechanisms to simplify it and make it fit the Technical system. Language stopped being about *"dreams, inspirations, aspirations and deliriums"* in the Technical milieu. Language is mathematicized[29], i.e. adapted to the modalities of Technique, it becomes subservient and creates a subservient version of Man.[30] *"Technology dissects in order to reconstruct, separates the elements of man in order to synthesise a man such as we have never known."*[31]. By this normative modelization, which is dissociation from our experience of the sensible world[32], it is submitting Man to the technical system that is seeked. Therefore Technique in all its applications masquerades as a rational process, that will forbid the spontaneous and irrational, with the ambition of modeling human brains after the computer : it does not leave space for human emancipation and autonomy, because they fundamentally contradict the capitalist structures sustaining economic growth.[33]

It is everything in language that is reduced to the mere "logical" and rationalisation, in the sense that everything becomes means and instruments for "efficiency" (growth)[34]. It is the mechanism of the assembly line and production norms that are the epitome of technical rationalizing to complete maximum efficiency. Another example we can give is transcription services, common across many domains and built into services such as Zoom and Google Meet. These services such as Otter AI[35] started out as essential tools for the inclusion of the deaf participant but now provide a much more detailed and sanitised verbalisation of the conversation. These tools aim to flatten verbalised language into simplistic and searchable terms.

"Efficiency" is a characteristic of Technique in Ellul's philosophical system, and that concept highlights the intricate economic imperatives behind technical rationality and therefore its technical language : it is dictated by the quest for what Marx called "capital accumulation", that equates human beings to costs that can be reduced by machinery or surplus value that can be extracted from their labour. Reducing Man to economic value was the only way to make things *"perfectly comprehensible and masterable"*[36] for the system to grow. (Ellul),

In recent years there has been a drive towards the creation of software to replace the role of the interpreter. Such works are being integrated across domains such as in work or education. Signapse AI[37] is a great example of that ambition. This work whilst technologically advanced provides uncanny real time avatars instead of the human based interpretation the deaf community needs and wants. The avatar, though superficially accurate in its reproduction of individual signs, often fails to capture the fluidity, timing, and embodied expressiveness that characterise natural signed interaction. Subtle elements such as facial grammar, eye gaze, spatial referencing, and affective nuance are either flattened or rendered inconsistently. The result is a form of communication that produces a dissonance between visual form and lived linguistic experience.Such work arises as instead of funding human interpreters, public services, community-based solutions, companies invest in AI tools because they are scalable, commodifiable, monetizable, on the promise that it will reduce costs and allow margins for profit and capital growth.

---

[29] André Gorz, *The Immaterial*, trans. Chris Turner (London: Seagull Books, 2010). p95
[30] Jacques Ellul, *Le système technicien* (Paris: Calmann-Lévy, 1977), p55
[31] Jacques Ellul, *La technique ou l'enjeu du siècle* (Paris: Economica, 2008).P393
[32] André Gorz, *The Immaterial*, trans. Chris Turner (London: Seagull Books, 2010).p95
[33] Jacques Ellul, *Le système technicien* (Paris: Calmann-Lévy, 1977), p 273
[34] Jacques Ellul, *La technique ou l'enjeu du siècle* (Paris: Economica, 2008). P86
[35] Otter.ai, "AI-Powered Real-Time Transcription," https://otter.ai/
[36] Jacques Ellul, *Le bluff technologique* (Paris: Hachette, 1988). p196

The other side of the coin is that marginalized communities need now to become new data mines to exploit, language and bodily expression are treated as raw data, and under platform capitalism, data is privatized, value is extracted without democratic control or compensation. If these AI tools reach radical monopoly, their imposition will constitute a continuity in coerced labour, which is a condition of capitalism, but a major obstacle to autonomy.

Technological growth is thus not only an ideological, moral and ontological project. As always in the history of Technique, the AI tools for deaf people do not aim to make them integrated socially but to serve the productivist system's given material interests. With accessibility reframed as a whole new market for an individualized, privately owned and merely technical solution to a structural problem, it is accessibility and inclusivity that are diluted in rampant exploitation, commodification, systemic injustice ; and because the inherent productivism of Technique operates on limitless expansion on a finite earth, degrowth philosophers such as Illich and Ellul pioneer the conclusion of inexorable environmental degradation that threatens the same and only ecosystem that can sustain human life itself.

---

[37] Signapse AI, "Sign Language Avatars for Business," https://www.signapse.ai/

### 6. <u>Reclaiming sign language</u>

But to Ivan Illich, political consciousness and praxis need to step away from the technical language and rationality to formulate democratic goals that may allow change. The "convivial" function of language needs to be recovered along with its imaginary, to regain community, autonomy and p*oiesis*.[38] : *"But the ability to direct events at that moment depends on how well these minorities grasp the profound nature of the crisis, and know how to state it in effective language: to declare what they want, what they can do, and what they do not need. The critical use of ordinary language is the first pivot in a political inversion."*[39]

When one begins to reclaim language, we do so as an act of liberation. We hope to restore the positivity of dialogue: removed from the constraints of efficiency speed or algorithmic normativity. The machine alone, no matter how sophisticated, cannot reproduce the subtle interweaving of memory, affect, and imagination through which meaning is lived and renewed. Here now lies the question of a correct approach - is it possible to map the human experience and linguistic diversity to the standardising gaze of the machine? Can we liberate the signed deaf experience from such normative constraints? Through certain methods such desired outputs may be achieved, although not without significant barriers.

Firstly we must begin to see signed language as an instrument of education and a repository of culture; rich with history and identity. This forms a particular touch point for hearing communities and those who actively research in such areas. It is rather 'easy' to design such AI recognition systems with the Deaf community as the end goal but it becomes rather unusual to allow them to be part of the process of creation itself. If we deign to argue that a system truly reflects the experience of the Deaf participant then we must simply include the desired parties in its design; actively.

The simplest way to see this in action is to look at the process of creation for any popular modern AI sign language solution. The starting point for any major project currently hinges on large scale, quality data. Within the UK this does form its own problem - with fewer public datasets available, such systems are reliant on out of date linguistic data. British Sign Language (BSL) is also a great example of such signed language that has great difference in execution across varied individuals within the country. Previously this issue was tackled in datasets such as the BSL Corpus[40], where the data itself was a recording of informal conversations between individuals across different regions - unlike its most popular successor BOBSL[41] which only consists of hearing interpreters signing formalised BSL for public broadcasting. This sudden change of linguistic focus has likely evolved from the straightforward and accessible nature of interpreted data rather than signed conversational data. Whilst this provides the speed and the sanitisation that the machine desires it however does not provide the human basis that the deaf community needs. This issue is further exasperated as datasets like BSL Corpus have remained untouched for many years and now consist of what we might refer to as *outdated linguistic data.* Therefore we must continuously collect, dissect and make use of data that reflects the real communicative acts of the deaf community.

---

[38] Ivan Illich, *Tools for Conviviality* (London: Fontana/Collins, 1973).p106
[39] Ivan Illich, *Tools for Conviviality* (London: Fontana/Collins, 1973). p120
[40] British Sign Language Corpus Project (London: University College London, 2013)
[41] "BOBSL: BBC-Oxford British Sign Language Dataset (2021)

Now that we understand the foundational technical problem, one must begin to recognise that the real life problems Deaf people face that such technologies aim to solve are not rooted in their bodies or minds, but in the refusal of hearing society to adapt, to learn sign language, or to embed accessible communication practices into everyday life. The problem is not the Deaf person's capacity to communicate, but the structural and social choices that prioritise convenience, efficiency, and hearing norms over inclusivity. To present AI or other technological systems as a "solution" to these barriers is therefore inherently ableist: it positions Deaf people as the problem to be fixed, rather than addressing the societal and institutional failures that create exclusion in the first place.

This dynamic is further compounded when such technologies are designed without the input or leadership of Deaf communities, producing tools that are not only insufficient but dehumanising. The act of reclaiming language, in this context, is also an act of resistance: it refuses the narrative that inclusivity can be outsourced to machines, insisting instead on a world in which communication, culture, and consciousness are shaped collaboratively and equitably. Through this emerges a broader alignment with contemporary critiques of technological solutionism, visible in growing movements of resistance that challenge the unchecked expansion of artificial intelligence. These include organised protests, labour actions, and forms of unionisation that contest the displacement of human agency, the extraction of cultural and linguistic labour, and the consolidation of power within technical systems and their governing institutions. Such AI (what we argue here as *ableist intelligence*), need not exist in a society that provides the necessary adaptations, where inclusion is embedded structurally rather than imposed through the disciplining of the Deaf minority. Real inclusivity, one that serves autonomy, justice, emancipation can only come with political, systemic and structural change and not more technology.

We do not, however, exist within such a perfect society. The expectation that people, as a collective will suddenly change through sheer advocacy alone is a farce. Now one begins to think like the hearing person designing the 'solution'. How do we solve this problem that none dare solve through the simplest manners? Where we do not feel some pressure to become more accessible? Here we can begin to understand that if addressed *with* the deaf collective rather than on *behalf of* , that AI may prove the bridge needed to close the communication gap provided by such societies. If we cannot mandate man to learn to sign then it falls on the machine to bridge this linguistic gap. Therefore the ableist nature of this intelligence will forever hinge on its execution not its sheer existence. Crucially, techno-solutionist approaches tend to frame the relationship between signed and spoken languages as a problem of transliteration rather than translation, reducing signed languages to sets of gestures to be mapped onto pre-existing linguistic structures. This misrecognition positions signed languages not as fully autonomous linguistic systems, but as derivative or embodied variants of spoken language. It is here that ableism is most deeply embedded: in the reduction of linguistic and cultural difference into forms that can be standardised, optimised, and rendered legible to the machine, rather than engaged with on their own terms.

## 7. **Conclusion**

This analysis has established that AI translation tools for the Deaf have transitioned from mere instruments into a dominant "milieu" that disciplines the user and orders the social world. We have demonstrated that these tools operate on a reductionist normativity, stripping sign language of its embodied, cultural, and social complexity. This leads to a colonization of consciousness, where language is standardized and individual subjectivity is alienated to fit technical imperatives. Consequently, these tools become counter-productive; by framing accessibility as a technical fix, they depoliticize inclusion and remove the social responsibility of the hearing majority to adapt. Finally, we highlighted how these systems function as an expression of productivism, prioritizing capital growth and data extraction over authentic human connection.

The central question originally asked whether AI tools designed for inclusivity might instead grow exclusion and be inherently ableist. After our analysis, the problem appears as follows: Can AI translation systems account for the fluid, intersubjective meaning-making of sign language without falling into a "radical monopoly" that standardizes and devalues Deaf culture for the sake of efficiency ?

Through the lens of degrowth/political ecology, we exposed the contradiction between technical solutions for "inclusivity" and our maximalist definition of inclusivity as autonomy, emancipation and flourishing. In exposing the counter-productivity of a seemingly ethically motivated AI product, we also ask if AI growth can ever serve "social good" at all and we challenge the widely spread alignment with a techno-solutionism that not only perpetuates harms, alienation, injustice, exploitation, environmental degradation, but will further grow and intensify these problems instead of solving them.

This critique does not inherently imply that all technological interventions or 'solutions' aimed at the subject of accessibility are ableist. Not every use of a particular technology participates equally within our technological ecosystem. Some technologies that remain subordinate to human agency and such are also developed with the Deaf community itself are able to resist such dynamics presented within our work here. Furthermore within other accessibility focused domains we may see AI as a far better tool for accessibility than its deaf counterpart. Such technologies include AI-powered smart glasses such as those pioneered by Envision[42] as well as assistive mobile applications such as 'Be My Eyes'.[43] The latter is a unique case, where the application is designed to function as a human-human support system, connecting its visually impaired users with sighted volunteers. Here AI has been integrated into an already existing ecosystem, giving its visually impaired users choice between AI functionality and human based support. Here it preserves the primacy of user control and contextual judgment, illustrating how AI can operate as an aid within communicative practices rather than as a totalising technical solution.

---

[42] Envision, https://www.letsenvision.com
[43] Be My Eyes, https://www.bemyeyes.com

After our very critical account, it is good to remember that according to Ellul, there's no possible reversal of technological development ; but regulation and responsibility for more deeply ethical AI can not be entrusted to companies, nor to States, which have lost sovereignty to monopolies and economic dependence. The responsibility for change therefore lies on individuals' ability to collectively organise, which is made increasingly difficult by the subjective and social alienation.caused by growing technological development. It seems that realizing autonomy requires autonomy : leaving us with a remaining difficulty, and very little hope for the future.

This analysis leads us to believe it is not possible to settle this argument of whether AI solutions for the accessibility of sign language users are, in principle, incapable of supporting deaf autonomy. Rather one must argue that instead it is best to conclude that within the current technological ecosystem (focused on structured efficiency, scalability and profit) that such systems tend towards an ableist standardisation. The critique therefore targets the *conditions of possibility* under which such technological solutions are developed and deployed, not every conceivable technical intervention involving computation and accessibility.

Therefore, we argue that further work will be required. Empirically there is a need for deaf-led participatory research in order to examine how existing AI translation systems are experienced, resisted and adapted by Deaf users in everyday contexts. Politically future research must interrogate whether forms of collective organisation can meaningfully intervene in the technological ecosystem or whether reverting to human based accessibility approaches is deemed more suitable and appropriate.
.

**Acknowledgements**

We would like to thank Tom Stoneham, Jennifer Chubb, Kofi Appiah, Darren Reed and the SAINTS CDT of University of York for the support, advice, and feedback received for this paper.